\pdfoutput=1
\documentclass[11pt,a4paper]{article}
\usepackage[hyperref]{emnlp2020}
\usepackage{times}
\usepackage{latexsym}
\usepackage{amsmath}
\usepackage{amsfonts}
\usepackage{amssymb}
\usepackage{latexsym}
\usepackage{booktabs}
\usepackage{color} 
\usepackage{setspace}
\usepackage{multirow}
\usepackage{url}
\usepackage{graphicx} 
\usepackage{subfigure}
\usepackage{algorithm}
\usepackage{algorithmic}
\usepackage{microtype}

\aclfinalcopy % Uncomment this line for the final submission
\title{LIMIT-BERT : Linguistics Informed Multi-Task BERT}

\author{Junru Zhou$^{1,2,3}$ , Zhuosheng Zhang $^{1,2,3}$,  Hai Zhao$^{1,2,3}$\thanks{$\ $  Corresponding author. This paper was partially supported by National Key Research and Development Program
of China (No. 2017YFB0304100), Key Projects of National Natural Science Foundation of China (U1836222 and
61733011), Huawei-SJTU long term AI project, Cutting-edge
Machine reading comprehension and language model.} , Shuailiang Zhang $^{1,2,3}$\\

$^{1}$Department of Computer Science and Engineering, Shanghai Jiao Tong University \\
	$^{2}$Key Laboratory of Shanghai Education Commission for Intelligent Interaction \\ and Cognitive Engineering, Shanghai Jiao Tong University, Shanghai, China\\
	$^{3}$MoE Key Lab of Artificial Intelligence, AI Institute, Shanghai Jiao Tong University \\
  {\tt \{zhoujunru,zhangzs\}@sjtu.edu.cn, zhaohai@cs.sjtu.edu.cn}}

\date{}

\begin{document}
\maketitle

\begin{abstract}

In this paper, we present Linguistics Informed Multi-Task BERT (LIMIT-BERT) for learning language representations across multiple linguistics tasks by Multi-Task Learning.
LIMIT-BERT includes five key linguistics tasks:
Part-Of-Speech (POS) tags, constituent and dependency syntactic parsing, span and dependency semantic role labeling (SRL).
Different from recent Multi-Task Deep Neural Networks (MT-DNN), our LIMIT-BERT is fully linguistics motivated and thus is capable of adopting an improved masked training objective according to syntactic and semantic constituents. Besides, LIMIT-BERT takes a semi-supervised learning strategy to offer the same large amount of linguistics task data as that for the language model training.
As a result, LIMIT-BERT not only improves linguistics tasks performance, but also benefits from a regularization effect and linguistics information that leads to more general representations to help adapt to new tasks and domains.
% LIMIT-BERT obtains new state-of-the-art or competitive results on both span and dependency semantic parsing and both dependency and constituent syntactic parsing. 
LIMIT-BERT outperforms the strong baseline Whole Word Masking BERT on both dependency and constituent syntactic/semantic parsing, GLUE benchmark, and SNLI task.
Our practice on the proposed LIMIT-BERT also enables us to release a well pre-trained model for multi-purpose of natural language processing tasks once for all.
% and eight out of
% nine GLUE tasks, pushing the GLUE benchmark
% to 82.7\%.
% The code
% and pre-trained models are publicly available
% at https://github.com/namisan/mt-dnn.

% 1. We present Linguistics Informed Multi-Task BERT (LIMIT-BERT) for learning language representations across multiple linguistics tasks by multi-task learning and semi-supervised learning.
% 2. We propose Linguistics-Guided Mask Strategy which intends to choose and predict meaningful words or phrases such as verb predicates or noun phrases to improve language model performance.
% 3. LIMIT-BERT outperforms the strong baseline Whole Word Masking BERT on both span and dependency semantic parsing and both dependency and constituent syntactic parsing and obtains new state-of-the-art results on SNLI tasks.

\end{abstract}

\section{Introduction}
Recently, pre-trained language models have shown greatly effective across a range of linguistics inspired natural language processing (NLP) tasks such as syntactic parsing, semantic parsing and so on \cite{zhou-zhao-2019-head, zhou2019parsing, ouchi-etal-2018-span, he-etal-2018-syntax, Li-aaai-19}, when taking the latter as downstream tasks for the former. In the meantime, introducing linguistic clues such as syntax and semantics into the pre-trained language models may furthermore enhance other downstream tasks such as various Natural Language Understanding (NLU) tasks \cite{zhang2019semanticsaware, zhang2019sgnet}. However, nearly all existing language models are usually trained on large amounts of unlabeled text data \cite{PetersN18-1202,Jacobbert}, without explicitly exploiting linguistic knowledge.
Such observations motivate us to jointly consider both types of tasks, pre-training language models, and solving linguistics inspired NLP problems. We argue such a treatment may benefit from two-fold. (1) Joint learning is a better way to let the former help the latter in a bidirectional mode, rather than in a unidirectional mode, taking the latter as downstream tasks of the former. (2) Naturally empowered by linguistic clues from joint learning, pre-trained language models will be more powerful for enhancing downstream tasks. Thus we propose Linguistics Informed Multi-Task BERT (LIMIT-BERT), making an attempt to incorporate linguistic knowledge into pre-training language representation models.
% LIMIT-BERT is based on the model proposed in \cite{zhou2019parsing} which is joint learning of 
% Besides, Multi-Task Learning (MTL) \cite{Caruana1993Multitask} has shown it is useful for multiple (related)
% tasks to be learned jointly so that the knowledge
% learned in one task can benefit other tasks which profits from a regularization effect via alleviating
% overfitting to a specific task, thus making
% the learned representations universal across tasks.
The proposed LIMIT-BERT is implemented in terms of Multi-Task Learning (MTL) \cite{Caruana1993Multitask} which has shown useful, by
% tasks. Its advantage comes from two sides. Firstly, the knowledge learned in one task can benefit other similar tasks inherently. Secondly, MTL shows a regularization effect via 
alleviating overfitting to a specific task, thus making the learned representations universal across tasks.
% Recent work (Strubell et al., 2018) has shown
% that for linguistic knowledge downstream semantic tasks with much
% smaller datasets, such as Semantic Role Labeling
% (SRL) (Palmer et al., 2005), self-attention models
% greatly benefit from the use of linguistic information
% such as dependency parsing annotations. Motivated
% by this work, we examine to what extent
% with required discourse and semantic annotations.
% we can use discourse and semantic information
% to extend self-attention-based neural models for a
% higher-level task such as Reading Comprehension.

Since universal language representations are learning by leveraging
large amounts of unlabeled data which has quite different data volume compared with linguistics tasks dataset such as Penn Treebank (PTB)\footnote{PTB is an English treebank with syntactic tree annotation which only contains 50k sentences.} \cite{MarcusJ93-2004}.

To alleviate such data unbalance on multi-task learning, we apply semi-supervised learning approach that uses a pre-trained linguistics model\footnote{The model may jointly predict syntax and semantics for both span and dependency annotation styles, which is from \cite{zhou2019parsing} and joint learning with POS tag.} to annotate large amounts of unlabeled text data and to combine with gold linguistics task dataset as our final training data.
For such pre-processing, it is easy to train our LIMIT-BERT on large amounts of data with many tasks concurrently by simply summing up all the concerned losses together.
Moreover, since every sentence has labeled with predicted syntax and semantics, we can furthermore improve the masked training objective by fully exploiting the known syntactic or semantic constituents during the language model training process.
Unlike the previous work MT-DNN \cite{liu-etal-2019-multi} which only fine-tunes BERT on GLUE tasks, our LIMIT-BERT is trained on large amounts of data in a semi-supervised way and firmly supported by explicit linguistic clues.

%  For example, BERT is based
% on a multi-layer bidirectional Transformer, and is
% trained on plain text for masked word prediction
% and next sentence prediction tasks.

% MT-DNN obtains new state-of-the-art results on
% eight out of nine NLU tasks 2 used in the General
% Language Understanding Evaluation (GLUE)
% benchmark (Wang et al., 2018), pushing the GLUE
% benchmark score to 82.7%, amounting to 2.2% absolute
% improvement over BERT.We further extend
% the superiority of MT-DNN to the SNLI (Bowman
% et al., 2015a) and SciTail (Khot et al., 2018)
% tasks. The representations learned by MT-DNN
% allow domain adaptation with substantially fewer
% in-domain labels than the pre-trained BERT representations.
% For example, our adapted models
% achieve the accuracy of 91.6% on SNLI and 95.0%
% on SciTail, outperforming the previous state-ofthe-
% art performance by 1.5% and 6.7%, respectively.
% Even with only 0.1% or 1.0% of the original
% training data, the performance of MT-DNN on
% both SNLI and SciTail datasets is better than many
% existing models. All of these clearly demonstrate
% MT-DNN’s exceptional generalization capability
% via multi-task learning.

We verify the effectiveness and applicability
of LIMIT-BERT on Propbank semantic parsing \footnote{It is also called semantic role labeling (SRL) for the semantic parsing task over the Propbank.} in both span style (CoNLL-2005) \cite{carreras-marquez-2005-introduction} and dependency style, (CoNLL-2009) \cite{hajic-etal-2009-conll} and Penn Treebank (PTB) \cite{MarcusJ93-2004} for both constituent and dependency syntactic parsing.
% the General Language Understanding Evaluation (GLUE) benchmark \cite{wang2018glue} and Stanford Natural Language Inference (SNLI) \cite{bowman-etal-2015-large} for Nat-ural Language Understanding (NLU) tasks.
% Additionally, we introduce a spacial label for SRL and syntactic parsing to take advantage of the relationship between dependency and span structure formulization, which enables our model to use a single decoder to implement SRL and syntactic parsing as HPSG parsing.
Our empirical results show that semantics and syntax can indeed benefit the language representation model via multi-task learning and outperforms the strong baseline Whole Word Masking BERT (BERT$_\text{WWM}$). 
% and LIMIT-BERT reaches new state-of-the-art or competitive performance for all four tasks: span and dependency SRL, constituent and dependency syntactic parsing.
% LIMIT-BERT also obtains new state-of-the-art on SNLI.

\section{Tasks and Datasets}

LIMIT-BERT includes five types of downstream
tasks: Part-Of-Speech, constituent and dependency syntactic parsing, span and dependency semantic role labeling (SRL).

Both span (constituent) and dependency are two broadly-adopted annotation styles for either semantics or syntax, which have been well studied and discussed from both linguistic and computational perspectives \cite{chomsky1981lectures,Li-aaai-19}.  
 
Constituency parsing aims to build a constituency-based parse tree from a sentence that represents its syntactic structure according to a phrase structure grammar.
While dependency parsing identifies syntactic relations
(such as an adjective modifying a noun) between
word pairs in a sentence. 
The constituent structure is better at disclosing phrasal continuity, while the dependency structure is better at indicating dependency relation among words.

Semantic role labeling (SRL) is dedicated
to recognizing the predicate-argument
structure of a sentence, such as \textbf{who} did
\textbf{what} to \textbf{whom}, where and when, etc.
For argument annotation, there are two formulizations.
One is based on text spans, namely span-based
SRL. The other is dependency-based SRL, which annotates
the syntactic head of argument rather than the entire argument
span.
SRL is an important method to obtain semantic information beneficial to a wide range of NLP tasks \cite{zhang2019explicit,mihaylov2019discourseaware}.

BERT is typically trained on quite large unlabeled text datasets,  BooksCorpus and English Wikipedia, which have 13GB plain text, while the datasets for specific linguistics tasks are less than 100MB.
Thus we employ semi-supervised learning to alleviate such data unbalance on multi-task learning by using a pre-trained linguistics model to label BooksCorpus and English Wikipedia data.
The pre-trained model jointly learns POS tags and the four types of structures on semantics and syntax, in which the latter is from the XLNet version of \cite{zhou2019parsing}, giving state-of-the-art or comparable performance for the concerned four parsing tasks.
During training, we set 10\% probability to use gold syntactic parsing and SRL data: Penn Treebank (PTB) \cite{MarcusJ93-2004},
span style SRL (CoNLL-2005) \cite{carreras-marquez-2005-introduction} and dependency style SRL (CoNLL-2009) \cite{hajic-etal-2009-conll}.

\label{Task and Dataset}

\subsection{Linguistics-Guided Mask Strategy}

\begin{figure}[t!]
    \centering
    \subfigure[Semantic Phrase Masking.]{
        \label{Fig.sub.1}
        \includegraphics[width=2.5in]{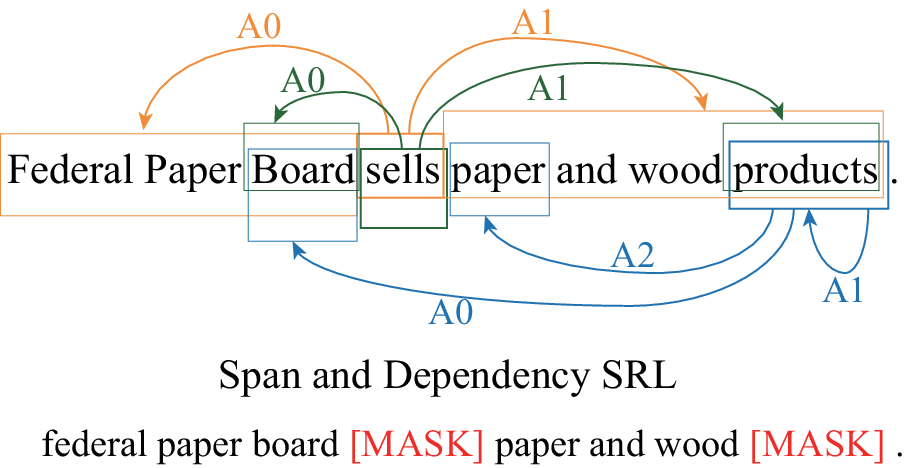}
    }
    \subfigure[Syntactic Phrase Masking.]{
        \label{Fig.sub.3}
        \includegraphics[width=2.5in]{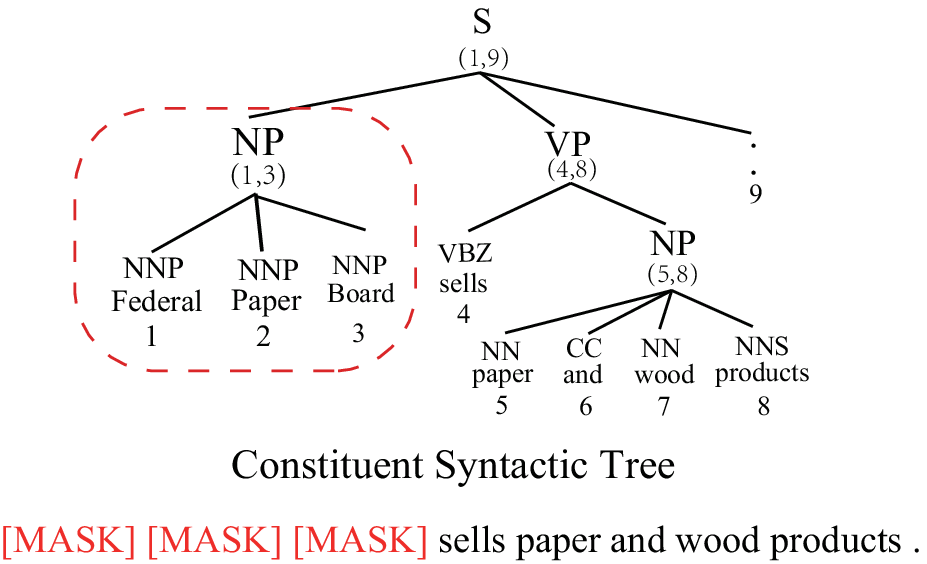}
    }
    \caption{Syntactic and Semantic Phrase Masking strategy. In figure (a) the predicates \textit{sells} and \textit{products} are replaced by [MASK] while in figure (b) each token of constituent \textit{federal paper board} also has been masked.}
    \label{fig1}
\end{figure}

BERT applies two training objectives:  Masked Language Model (LM) and  Next Sentence Prediction (NSP) based on  WordPiece embeddings \cite{wu2016googles} with a 30,000 token vocabulary. For Masked LM training objective, BERT uses training data generator to 
choose 15\% of the token positions at random for mask replacement and predict the masked tokens\footnote{Actually, BERT applies three replacement strategies: (1) the [MASK] token 80\% of
the time (2) random token 10\% of the time (3)
the unchanged $i$-th token 10\% of the time. This work uses the same replacement strategies.}.
Since using different masking strategy can improve model performance such as the Whole Word Masking\footnote{https://github.com/huggingface/transformers} which masks all of the tokens corresponding to a word at once, we further improve the masking strategy by exploiting available linguistic clues, syntactic or semantic constituents (phrases)\footnote{Syntactic phrases indicate the constituent subtrees while semantic phrases represent as predicate or argument in span SRL.}, predicted by our pre-trained linguistics model as discussed in Section \ref{Task and Dataset}.
Thus, we apply three mask strategies at random for each sentence: Syntactic Phrase Masking, Semantic Phrase Masking, and Whole Word Masking.
Syntactic/Semantic Phrase Masking (SPM) means that all the tokens corresponding to a syntactic/semantic phrase are masked, as shown in Figure \ref{fig1}.
The overall masking rate and replacement strategy remain the same as BERT, we still predict each masked WordPiece token independently.
Intuitively, it makes sense that SPM is strictly more powerful than original Token Masking or Whole Word Masking, since SPM may choose and predict the meaningful words or phrases such as verb predicates or noun phrases.

\begin{figure*}[t!]
    \centering
    \includegraphics[width=6.4in]{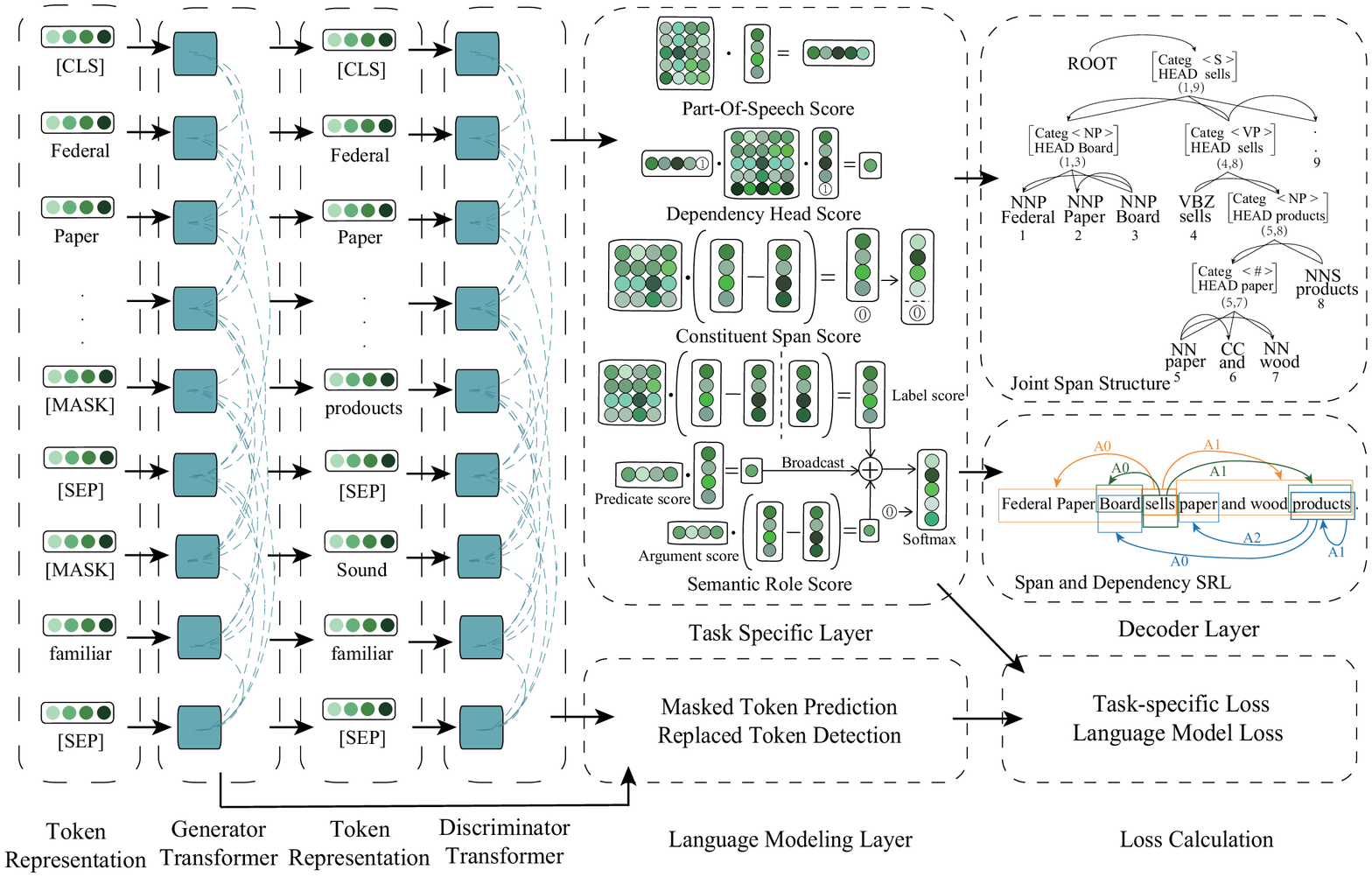}
    
    \caption{The framework of LIMIT-BERT.}
    \label{fig2}
\end{figure*}

\section{LIMIT-BERT Model}

\subsection{Overview}

The architecture of the LIMIT-BERT is shown
in Figure \ref{fig2}. Our model includes four modules: token representation, Transformer encoder, language modeling layers, task-specific layers including syntactic and semantic scorers and decoders.
We take multi-task learning (MTL) \cite{Caruana1993Multitask} sharing the parameters of token representation and Transformer encoder, while language modeling layers and the top task-specific layers have independent parameters.
The training procedure is simple that we just sum up the language model loss with task-specific losses together.
% The input X is based on WordPiece sequence and either a sentence or a pair of sentences packed
% together, while our linguistic tasks are based on word level of only one sentence.

\subsection{Token Representation}

Following BERT token representation \cite{Jacobbert}, the first token is always the [CLS] token. If input $X$ is packed by a sentence pair $X_1$; $X_2$, we separate the two sentences
with a special token [SEP] ("packed by" means connect two sentences as BERT training). The Transformer encoder
maps $X$ into a sequence of input embedding vectors,
one for each token, which is a sum of
the corresponding word, segment, and positional
embeddings.

If we apply BERT training data (BooksCorpus and English Wikipedia), we use pair sentences packed to perform next sentence prediction and only take the first sentence including [CLS] and [SEP] token for later linguistics tasks.
While using gold linguistics task data (PTB, CoNLL-2005, and CoNLL-2009) with 10\% probability, we only take one sentence as input that [CLS] and [SEP] are first and last tokens respectively.

Since input sequence $X$ is based on WordPiece token, we only take the last WordPiece vector of the word in the last layer of Transformer encoder as our sole word representation for later linguistics tasks input to keep the same length of the token and label annotations.

\label{Token Representation}

\subsection{Transformer Encoder}

The Transformer encoder in our model is adapted from \cite{Vaswani17}, which transforms the input representation vectors into a sequence of contextualized embedding vectors with shared representation across different tasks. We use the pre-trained parameters of BERT \cite{Jacobbert} as our encoder initialization for faster convergence.
% Unlike MT-DNN model \cite{liu-etal-2019-multi}, we also share the language model objective with multi-task learning on a large scale linguistic labeled data.
% using the NLU tasks in GLUE as examples,
% although in practice we can incorporate arbitrary
% natural language tasks such as text generation
% where the output layers are implemented as a neural
% decoder.
\label{Transformer Encoder}

\subsection{Language Modeling Layers}

BERT training applies masked language modeling (MLM) as a training objective which corrupts the input by replacing some
tokens with a special token [MASK] and then lets the model reconstruct the original tokens.
While in our LIMIT-BERT training, the linguistics specific tasks and MLM training take the same input; thus the [MASK] tokens raise a mismatch problem that the model sees artificial [MASK] tokens during MLM training but
not when being fine-tuned and inference on linguistics tasks.
Besides, due to learning bidirectional
representations, MLM approaches incur a substantial computational cost increase because the network only learns from 15\% of the tokens per example and needs more training time to converge.

Recently \cite{XLNet-Zhilin-2019,anonymous2020electra} have made attempts to alleviate such a difficulty. The latter applies a replaced token detection task in their ELECTRA model.
Instead of masking the input, ELECTRA corrupts it
by replacing some input tokens with plausible alternatives sampled from a small generator network, which is close to the original input without [MASK] tokens.

We adopt the ELECTRA training approach in our LIMIT-BERT, which lets the generator $G$ and discriminator $D$ share the same parameters and embedding as shown in Figure \ref{fig2}.
The generator $G$ is identical to BERT training \cite{Jacobbert} that predicts the masked tokens and next sentence and sums token mask loss and next sentence predict loss\footnote{If using gold linguistics task data, we only compute the token mask loss.} as $J_{G}(\theta)$.
Then the discriminator $D$ takes the predicted tokens by generator $G$\footnote{For the non-masked tokens, we take the original tokens as input.} and is trained to distinguish tokens that have been replaced by generator $G$ which is a simple binary classification of each token with loss $J_{D}(\theta)$.
At last, we take the output vector $X$ of discriminator $D$ to feed the following task-specific layers and sum the loss of $J_{G}(\theta)$ and $J_{D}(\theta)$ as the final language modeling loss $J_{lm}(\theta)$:
$$
J_{lm}(\theta) =  J_{G}(\theta) +\lambda J_{D}(\theta),
$$
\noindent where $\lambda$ is set to 50 as the same as ELECTRA.
\label{Language Modeling Layers}
\subsection{Task-specific Layers}

Firstly, we rebuild word representations from the WordPiece tokens for linguistics tasks.
Then we follow \cite{zhou2019parsing} to construct the task-specific layers, including scoring layer and decoder layer.
The former scores three types of linguistic objectives, dependency head, syntactic constituent and semantic role.
The latter is to generate the legal linguistics structures.

\noindent \textbf{Word Level Construction} \quad
Suppose that $X$ is the output of the discriminator Transformer encoder, we pre-process the WordPiece sequence vector $X$ for linguistics specific tasks learning which are based on word level.
We only take the first sentence $X_1$ including token [CLS] and [SEP] of packed sentence pair ($X_1$; $X_2$).
Then we convert WordPiece sequence vector to word-level by simply taking the last WordPiece token vector of the word as the representation of the whole word.

\noindent \textbf{Scoring Layer} \quad

After the word-level construction, we calculate the POS tag, syntactic constituent,
 dependency head, and semantic role scores, 
%  following the way as \cite{zhou2019parsing} to construct span representation $s_{ij}$ for both syntactic constituent and semantic role scoring.
 following the training way as \cite{zhou2019parsing} to construct syntactic constituent, dependency head, and semantic role scores objective loss which are represented as $J_1(\theta)$, $J_2(\theta)$ and $J_3(\theta)$ respectively.

For POS tagging model training, we apply a one-layer feedforward network and minimize the negative log-likelihood of the gold POS tag $gp_i$ of each word, which is implemented as a cross-entropy loss:
$$
J_4(\theta) = -logP_{\theta}(gp_i|x_i),
$$
\noindent where $x_i$ is word vector inside $X$.

Utilizing these specific task scores, we do a sum to obtain the linguistics task loss $J_{lt}(\theta)$ for training:
$$
J_{lt}(\theta) = J_1(\theta) + J_2(\theta) + J_3(\theta) + J_4(\theta).
$$
At last, our LIMIT-BERT is trained for simply minimizing the overall loss:
$$
J_{overall}(\theta) = J_{lm}(\theta) +J_{lt}(\theta).
$$
\noindent \textbf{Decoder Layer} \quad
For syntactic parsing, we apply the $joint$ $span$ CKY-style algorithm to generate constituent and dependency syntactic tree simultaneously by following \cite{zhou-zhao-2019-head}.
% \footnote{Besides, for constructing a full predicted syntactic tree, we also  join POS tasks in our model and use POS score to predict the POS tags.}.
 
 For span and dependency SRL, we use a single dynamic programming decoder according to the uniform semantic role score following the non-overlapping constraints: span semantic arguments for the same predicate do not overlap \cite{punyakanok-2008-importance}.
For further details of the scoring and decoder layer, please refer to \cite{zhou2019parsing}.

\section{Experiments}

% We evaluate the proposed MT-DNN on three popular
% NLU benchmarks: GLUE (Wang et al., 2018),
% SNLI (Bowman et al., 2015b), and SciTail (Khot
% et al., 2018). We compare MT-DNN with existing
% state-of-the-art models including BERT and
% demonstrate the effectiveness of MTL with and
% without model fine-tuning using GLUE and domain
% adaptation using both SNLI and SciTail.

\subsection{Evaluation}

We use the model of \cite{zhou2019parsing} with fine-tuned uncased BERT$_\text{WWM}$ (whole word masking) as the baseline\footnote{Our codes and the pre-trained models : https://github.com/DoodleJZ/LIMIT-BERT.}.
For fairly compared to the baseline BERT$_\text{WWM}$, we also extract the language modeling layer of LIMIT-BERT and use the same model of \cite{zhou2019parsing} to fine-tune.
We evaluate our proposed model LIMIT-BERT 
and baseline model BERT$_\text{WWM}$ on CoNLL-2009 shared task \cite{hajic-etal-2009-conll} for dependency-style SRL, CoNLL-2005 shared task \cite{carreras-marquez-2005-introduction} for span-style SRL both using the Propbank convention \cite{palmer-etal-2005-proposition}, and English Penn Treebank (PTB) \cite{MarcusJ93-2004} for constituent syntactic parsing, Stanford basic dependencies (SD) representation \cite{Marieffe06generatingtyped} converted by the Stanford parser{\footnote{http://nlp.stanford.edu/software/lex-parser.html}} for dependency syntactic parsing using the same model of \cite{zhou2019parsing} to fine-tune.
We follow standard data splitting and evaluate setting as \cite{zhou2019parsing} and use end-to-end SRL setups of both span and dependency SRL.
Since LIMIT-BERT involves all syntactic and semantic  parsing  tasks, it is possible to directly apply LIMIT-BERT to each task without fine-tuning and we also compare these results.

In order to evaluate the language model pre-training performance of our LIMIT-BERT, we also evaluate LIMIT-BERT on two widely-used datasets, The General Language Understanding Evaluation (GLUE) benchmark \cite{wang2018glue} which is a collection of nine NLU tasks and Stanford Natural Language Inference (SNLI) \cite{bowman-etal-2015-large} to show the superiority.

% For the predicate disambiguation task in
% dependency SRL, we follow \cite{marcheggiani-titov-2017-encoding} and use the off-the-shelf disambiguator
% from \cite{roth-lapata-2016-neural}. 
% For constituent syntactic parsing, we use the standard evalb{\footnote{http://nlp.cs.nyu.edu/evalb/}} tool to evaluate the F1 score. For dependency syntactic parsing, following previous work \cite{Dozat2017Deep}, we report the results without punctuations of the labeled and unlabeled attachment
% scores (LAS, UAS).

\subsection{Implementation Details}

Our implementation of LIMIT-BERT is based on
the PyTorch implementation of BERT\footnote{https://github.com/huggingface/pytorch-pretrained-
BERT. We use Whole Word Masking BERT parameters
as our Transformer encoder initialization.}. 
We use a learning rate of 3e-5 and a batch size of 32 with 1 million training steps.
The optimizer and other training settings are the same as BERT \cite{Jacobbert}.
For task-specific layers including syntactic and semantic scorers and decoders, we set  the same hyperparameters settings as \cite{zhou2019parsing}.
LIMIT-BERT model is trained on 32 NVIDIA GeForce GTX 1080Ti GPUs. 
% four NVIDIA Titan RTX GPU 
% with Intel i7-7800X CPU of 20 days. 

% \noindent \textbf{Training Details}\quad we use 0.33 dropout for biaffine attention and MLP layers. All models are trained for up to 150 epochs with batch size 150 on a single NVIDIA GeForce GTX 1080Ti GPU with Intel i7-7800X CPU. 
% We use the same training settings as \cite{Kitaev-2018-SelfAttentive} and \cite{kitaev2018multilingual}.

% \subsection{Span Representation}

% As mentions in section 
% Different token representation combinations are evaluated in Table \ref{table1}. We find that CharLSTM performs a little better than CharCNNs. Moreover, POS tags on parsing performance show that predicted POS tags decreases parsing accuracy, especially without word information. 
% If POS tags are replaced by word embeddings, the performance increases.
% Finally, we apply word and CharLSTM as token representation setting for our full model{\footnote{We also evaluate POS tags on CTB which increases parsing accuracy, thus we employ the word, POS tags and CharLSTM as token representation setting for CTB.}}.

\subsection{Main Results}

\noindent \textbf{Syntactic Parsing Results} \quad
% Compared to the existing state-of-the-art models with pre-training, our LIMIT-BERT achieves 95.84 F1 score of constituent parsing and 97.14\% UAS and 95.44\% LAS of dependency parsing.
As shown in Table \ref{syndep} and \ref{synconst}, LIMIT-BERT
% outperforms all the published state-of-the-art models and 
without fine-tuning obtains 95.84 F1 score of constituent parsing and 97.14\% UAS and 95.44\% LAS of dependency parsing.
% and achieves the state-of-the-art performance of all the published models.
Compared with baseline BERT$_\text{WWM}$, LIMIT-BERT outperforms the baseline model both of fine-tuning or not.
Particularly, LIMIT-BERT without fine-tuning exceeds more than 0.2 in UAS of dependency and 0.1 F1 of constituent syntactic parsing which are considerable improvements on such strong baselines.

% our model and our model achieves new state-of-the-art on both constituent and dependency syntactic parsing without pre-training.

% Syntactic Parsing Results illustrate
% Compared with \cite{strubell-etal-2018-linguistically} shows that our joint model setting boosts both of syntactic parsing and SRL which are consistent with \cite{shi-etal-2016-exploiting} that syntactic parsing and SRL benefit relatively more from each other.

% In addition, we first report constituent parsing result of our models on
% the CoNLL-2005 Brown test set.

\begin{table}[t!]
	\begin{center}
	\small
	\resizebox{\linewidth}{!}{	
		\begin{tabular}{lcc}
			\hline
% 			\multirow{2}{*}{} & \multicolumn{2}{c}{WSJ} \\
% 			\cline{2-3}
			\multirow{2}{*}&UAS &LAS \\
			\hline
			%\cite{MaI17-1007} &94.88 &92.98\\
			\citet{Dozat2017Deep} &95.74 &94.08 \\
			\citet{Ma2018Stack}  &95.87 &94.19 \\
% 			\citet{strubell-etal-2018-linguistically}&94.92 &91.87  \\
            \citet{ji-etal-2019-graph} &95.97 &94.31 \\
			\citet{Daniel-2019-naacl-left}&96.04 &94.43 \\
			\citet{liu-etal-2019-hierarchical}& 96.09& 95.03\\
% 			\citet{zhou-zhao-2019-head} &96.09 &94.68 \\
% 			\citet{WangD18-1311}(ELMo) &96.35 &95.25  \\
% 			\citet{strubell-etal-2018-linguistically} &96.48 &94.40  \\
% 			\citet{zhou-zhao-2019-head}(ELMo) &96.76 &94.68 \\
			\citet{zhou-zhao-2019-head}(BERT) &97.00 &95.43 \\
% 			\citet{zhou-zhao-2019-head}(XLNet) &97.20 &95.72 \\
% 			\citet{zhou2019parsing}(ELMo)  &96.72 &95.00  \\
			\citet{zhou2019parsing}(BERT)  & 96.90 &95.32  \\
			\citet{zhou2019parsing}(XLNet)  & 97.23 &95.65  \\
			\hline
			Baseline (BERT$_\text{WWM}$) &96.89 &95.22  \\
			\bf Our LIMIT-BERT  &96.94 &95.30  \\
			\bf Our LIMIT-BERT$\dagger$  &97.14 &95.44  \\
			\hline
		\end{tabular}}
	\end{center}
	\caption{\label{syndep} 
% 	Comparison with published works of 
	Dependency syntactic parsing on PTB, no finetuning result is marked by $\dagger$.}
\end{table}

\begin{table}[t!]
	\centering
	\small
		\resizebox{\linewidth}{!}{
		\begin{tabular*}{\hsize}{@{}@{\extracolsep{\fill}}lccc@{}}
			\hline
			               &LR &LP &F1\\
			\hline 
            % \citet{SternP17}&93.2 &90.3 &91.8\\
			\citet{Gaddy} &91.76 &92.41 &92.08\\
% 			\citet{SternD17b} &92.57 &92.56 &92.56\\
% 			\citet{Kitaev-2018-SelfAttentive}  &93.20  &93.90 &93.55\\
% 			\citet{zhou-zhao-2019-head} &93.64 &93.92 &93.78 \\
% 			\hline
%             \bf Ours (wo/dep)     &93.56  &94.01 &93.79 \\
% % 			\bf Ours (w/dep)    &93.89  &94.10 &\bf94.00 \\
% 			\bf Ours (w/dep)    &93.94  &94.20 &\bf94.07 \\
% 			\hline
% 			\bf Pre-training \\
			\citet{Kitaev-2018-SelfAttentive}(ELMo) &94.85 &95.40 &95.13\\
			\citet{kitaev2018multilingual}(BERT) &95.46 &95.73 &95.59\\
 			% \citet{zhou-zhao-2019-head}(ELMo) &95.04 &95.39 &95.22 \\
 			\citet{zhou-zhao-2019-head}(BERT)&95.70 &95.98 &95.84 \\
 			% \citet{zhou-zhao-2019-head}(XLNet) &96.21 &96.46 &96.33 \\

            % \citet{zhou2019parsing}(ELMo)  &95.07  &95.40 &95.23  \\
			\citet{zhou2019parsing}(BERT)  &95.39  &95.64 &95.52  \\
			\citet{zhou2019parsing}(XLNet) &96.10  &96.26 &96.18  \\
			\hline
			Baseline (BERT$_\text{WWM}$) &95.59  &95.86 &95.72 \\
			\bf Our LIMIT-BERT  &95.67  &95.92 &95.80  \\
			\bf Our LIMIT-BERT$\dagger$  &95.72  &95.96 &95.84  \\
			\hline
		\end{tabular*}}
	\caption{\label{synconst} 
% 	Comparison with published works of 
	Constituent syntactic parsing on PTB, no finetuning result is marked by $\dagger$.}
\end{table}

\begin{table}[t!]
	\centering
	\resizebox{\linewidth}{!}{
	\begin{tabular}{lcccccc}  
		\toprule  
		\multirow{2}{*}{System}
		&\multicolumn{3}{c}{WSJ}
		&\multicolumn{3}{c}{Brown}
		\cr  
		\cmidrule(lr){2-4} 
		\cmidrule(lr){5-7} 
		 &P&R&F$_1$&P&R&F$_1$\cr
	\midrule
	\textit{End-to-end Span SRL} \cr
% 	\citet{he-etal-2017-deep} &80.2 &82.3 &81.2 &67.6 &69.6 &68.5\cr
    \citet{he-etal-2018-jointly} &81.2 &83.9 &82.5 &69.7 &71.9 &70.8\cr
    % \citet{Li-aaai-19}(ELMo) &- &- &83.0 &- &- &-\cr
    % \citet{Tan-Deep-Semantic}&84.5 &85.2 &84.8 &73.5 &74.6 &74.1 \cr
    % \citet{strubell-etal-2018-linguistically}
    % &84.07 &83.16 &83.61 &73.32 &70.56 &71.91\cr
    % \citet{strubell-etal-2018-linguistically}(ELMo)
    % &85.53 &84.45 &84.99 &75.8 &73.54 & 74.66\cr
    % \midrule
    % \bf Ours (wo/dep) &83.65&85.48&84.56&72.02&73.08&72.55\cr
    % \bf Ours (w/dep) &83.54&85.30&84.41&71.84&72.07&71.95\cr
    \citet{he-etal-2018-jointly}(ELMo) &84.8 &87.2 &86.0 &73.9 &78.4 &76.1\cr
    \citet{Li-aaai-19}(ELMo) &85.2 &87.5 &86.3 &74.7 &78.1 &76.4\cr
    % \citet{strubell-etal-2018-linguistically}
    % &86.69 &86.42 &86.55 &78.95 &77.17 &78.05\cr
    \citet{strubell-etal-2018-linguistically}(ELMo)
    &87.13 &86.67 &86.90 &79.02 &77.49 &78.25\cr
    % \citet{zhou2019parsing}(ELMo) &85.33&87.70&86.50&75.95&78.30&77.11\cr
    \citet{zhou2019parsing}(BERT) &86.46&88.23& 87.34&77.26&80.20&78.70\cr
    \citet{zhou2019parsing}(XLNet) &87.48&89.51&88.48&80.46&84.15&82.26\cr
    
   \midrule
   Baseline (BERT$_\text{WWM}$) 
%   &-&-&- &-&-&-\cr
    &86.48&88.59&87.52  &79.4&82.68&81.01\cr
    \bf Ours LIMIT-BERT
    &86.62&89.12&87.85&79.58&83.05&81.28\cr
     \bf Ours LIMIT-BERT$\dagger$
    &87.16&88.51&87.83&79.20&80.29&79.74\cr

	\midrule 
	\textit{End-to-end Dependency SRL} \cr
% 	\cite{zhao-2014-Integrative} &$-$ &$-$ &82.5 &$-$ &$-$ &$-$\cr
    \citet{Li-aaai-19} &- &- &85.1 &- &- &-\cr
    % \midrule
    % \bf Ours (wo/dep) &84.24&87.55&\bf 85.86&76.46&78.52&\bf 77.47\cr
    % \bf Ours (w/dep) &83.73&86.94& 85.30&76.21&77.89& 77.04\cr
    \citet{he-etal-2018-syntax} &83.9 &82.7 &83.3 &- &- &-\cr
    \citet{cai-etal-2018-full}&84.7 &85.2 &85.0 &- &- &72.5\cr
    \citet{Li-aaai-19}(ELMo) &84.5 &86.1 &85.3 &74.6 &73.8 &74.2\cr
    % \bf Ours LIMIT-BERT &85.21&88.17&86.66&78.62&80.76&79.68\cr
    % \citet{zhou2019parsing}(ELMo) &84.85&88.21& 86.50&78.43&80.52&79.46\cr
    % \bf Ours (wo/dep) + BERT &87.40&88.96&88.17&80.32&82.89&81.58\cr
    \citet{zhou2019parsing}(BERT) &86.77&89.14& 87.94&79.71&82.40& 81.03\cr
    % \bf Ours (wo/dep) + XLNet
    % &86.58&90.40&\bf88.44&80.96&85.31&\bf83.08\cr
    \citet{zhou2019parsing}(XLNet) &86.35&90.16&88.21&80.90&85.38&83.08\cr
    \midrule
    Baseline (BERT$_\text{WWM}$) 
    % &-&-&- &-&-&-\cr
    &85.13& 89.21&87.12 &79.05&83.95&81.43\cr
    
    \bf Ours LIMIT-BERT
    &85.84&90.01&87.87&79.50&84.85&82.09\cr
    \bf Ours LIMIT-BERT$\dagger$
    &85.73&89.34&87.50&79.60&82.81&81.17\cr
    
	\bottomrule  
	\end{tabular}
	}
	\caption{
% 	Comparison with published works of 
	Span SRL and dependency SRL results on on CoNLL-2005 and CoNLL-2009 test sets in end-to-end mode, no finetuning result is marked by $\dagger$.}\label{SRL results}
\end{table}

\noindent \textbf {Semantic Parsing Results} \quad
% We present all results on the CoNLL-2005 and CoNLL-2009 shared tasks in end-to-end mode.
Table \ref{SRL results} shows results on CoNLL-2005, CoNLL-2009 in-domain (WSJ) and out-domain (Brown) test sets and compares our LIMIT-BERT with previous published state-of-the-art models in end-to-end mode.
The upper part of the table presents results from span SRL while the lower part shows results of dependency SRL.
% Our LIMIT-BERT achieves new state-of-the-art in all SRL datasets  (both of in-domain and out-domain of span and dependency SRL).
% three SRL datasets  (in-domain and out-domain of span SRL and out-domain of dependency SRL) and competitive results of in-domain dependency SRL.
Compared with baseline, LIMIT-BERT with fine-tuning outperforms BERT$_\text{WWM}$ on all four SRL datasets, exceeding more than 0.3 in F1 of in-domain span SRL and 0.7 F1 of dependency SRL, which demonstrate that LIMIT-BERT can furthermore improve SRL performance even over strong baselines.

The results of syntactic and semantic parsing empirically illustrate that incorporating  linguistic  knowledge  into  pre-trained language model by multi-task and semi-supervised  learning  can significantly enhance downstream tasks.

% Our model outperforms the previous models with absolute improvements in F$_1$-score of 0.3\% on CoNLL-2005 benchmark. Besides, our single model performs even much better than all previous ensemble systems.
% %On all datasets, our model is able to predict over 40\% of the sentences completely correctly

%\paragraph{Dependency SRL}
% \noindent \textbf{Dependency SRL Results}\quad Table \ref{Dependency SRL results} presents the results on CoNLL-2009. We obtain new state-of-the-art both of end-to-end and given predicate mode and both of in-domain and out-domain text. 
% These results demonstrate that our improved uniform SRL representation can be adapted to perform dependency SRL and achieves impressive performance gains.
\noindent \textbf {SNLI Results}  \quad
Table \ref{tab:snli} includes the best results
reported in the leaderboards\footnote{\url{https://nlp.stanford.edu/projects/snli/}} of SNLI. We see that LIMIT-BERT outperforms the strong baseline model BERT$_\text{WWM}$ in 0.3 F1 score on the SNLI benchmark.
% best results
% reported in the leaderboards and even outperforms all the ensemble models\footnote{\url{https://nlp.stanford.edu/projects/snli/}. As ensemble models are commonly composed of multiple heterogeneous models and resources, we exclude them in our table to save space.} by a large margin. 

\noindent \textbf {GLUE Results} \quad
We fine-tuned LIMIT-BERT  for each GLUE task on
task-specific data.
The dev results in Table \ref{GLUE dev set results} show that LIMIT-BERT outperforms the strong baseline model and achieves remarkable results compared to other state-of-the-art models in literature. 

\begin{table}[t!]
		\resizebox{\linewidth}{!}
	{
	\begin{tabular}{p{5cm} p{1cm} p{1cm}}
		\hline
		
		\hline
		\textbf{Model} &  \textbf{Dev} & \textbf{Test}   \\ 
		\hline
% 		\multicolumn{3}{c}{\emph{In literature}} \\
		DRCN \cite{kim2018semantic} &- & 90.1\\
		SJRC \cite{zhang2019explicit} & - & 91.3 \\
		MT-DNN \cite{liu-etal-2019-multi} &92.2 & 91.6\\
		SemBERT \cite{zhang2019semanticsaware} & -  & 91.9\\
		\hline
% 		\multicolumn{3}{c}{\emph{Our implementation}} \\
		Baseline (BERT$_\text{WWM}$)  & 91.7  & 91.4  \\
		\textbf{LIMIT-BERT} & 92.3 & 91.7 \\
		\hline
		
		\hline
	\end{tabular}
}
	{
		\caption{\label{tab:snli} Leaderboards of SNLI dataset. Both our LIMIT-BERT and BERT$_\text{WWM}$ are single models.} } 
	
\end{table}

\begin{table*}
	\centering
	\resizebox{\linewidth}{!}
	{
		\begin{tabular}{lccccccccc}
			\hline
			
			\hline
% 			\textbf{Method} &  \multicolumn{2}{c}{\textbf{Classification}} &\multicolumn{3}{c}{\textbf{Natural Language Inference}} & \multicolumn{3}{c}{\textbf{Semantic Similarity}} &  \textbf{Score}\\
\textbf{Model} & CoLA & SST-2 & MRPC & STS-B & QQP  & MNLI  & QNLI & RTE & Score\\
%\textbf{Model} & CoLA & SST-2 & MNLI & QNLI & RTE  & MRPC  & QQP & STS-B & Score\\ 
&  (mc) & (acc)	& (F1/acc) & (pc/sc) & (acc/F1)  & m/mm(acc) & (acc) & (acc) & -\\
\hline
\multicolumn{10}{c}{\emph{Dev set results for Comparison}} \\
    BERT &60.6 &93.2 &-/88.0 &-/90.0 &91.3/- &-/86.6 &92.3 &70.4 &84.0 \\
% RoBERTa & 68.0 & 96.4 & -/90.2 & 94.7 & 86.6 & 90.9 & 92.2 & 92.4 & 88.9 \\
% XLNet & 63.6 & 95.6 & -/89.8 & 93.9 & 83.8 & 89.2 & 91.8 & 91.8 & 87.4 \\
% ELECTRA & 69.3 & 96.0 &  -/90.5 & 94.5 & 86.8  & 90.6 & 92.4  & 92.1  & 89.0\\
% MT-DNN & 63.5 & 94.3 & 87.1/86.7 & 92.9 & 83.4 & 87.5 & 89.2 & 90.6 & - \\
MT-DNN &63.5 &94.3 &91.0/87.5 &90.7/90.6 &91.9/89.2 &87.1/86.7 & 92.9 &83.4 &- \\
ELECTRA &69.3 &96.0 &-/90.6 &-/92.1 &92.4/- &-/90.5 &94.5 &86.8 &89.0 \\
\hline
% \multicolumn{10}{c}{\emph{Our implementation on dev set}} \\
Baseline (BERT$_\text{WWM}$)   & 63.6 & 93.6 & 90.8/87.0 & 90.5/90.2 &91.7/88.8 & 87.4/87.2 &93.9 & 77.3 &85.6 \\
\textbf{LIMIT-BERT} & 64.0 & 94.0 & 94.0/91.7 & 91.5/91.3 & 91.6/88.6 & 87.4/87.3 & 93.5 & 85.2 & 87.3 \\
\hline

\hline
\multicolumn{10}{c}{\emph{Test set results for models with standard single-task finetuning}} \\
% 			BiLSTM+ELMo+Attn &36.0 & 90.4 &76.4/76.1&79.9& 56.8 &84.9&64.8  & 75.1 &  70.5\\
BiLSTM+ELMo+Attn &36.0 &90.4 &84.9/77.9 &75.1/73.3 &84.7/64.8 &76.4/76.1 &- &56.8 &70.5\\
BERT &60.5 &94.9 &89.3/85.4 &87.6/86.5 &89.3/72.1 &86.7/85.9 &92.7 &70.1 &80.5\\
MT-DNN &62.5 &95.6 &91.1/88.2 &89.5/88.8 &89.6/72.7 &86.7/86.0 &93.1 &81.4 &82.7\\
SemBERT & 62.3 & 94.6 & 91.2/88.3 & 87.8/86.7 &89.8/72.8 & 87.6/86.3 & 94.6 & 84.5 & 82.9\\
% RoBERTa & 67.8 & 96.7 & 92.3/89.8 & 92.2/91.9 & 74.3/90.2 & 90.8/90.2 & 98.9 & 88.2 & 88.5 \\
ELECTRA & 71.7 & 97.1 & 93.1/90.7 & 92.9/92.5 & 90.8/75.6 & 91.3/90.8 & 95.8 & 89.8 & 89.35 \\

\textbf{LIMIT-BERT} & 62.5 & 94.5 & 90.9/88.0 & 90.3/89.7 & 89.5/71.9 & 87.1/86.2 & 94.0 & 83.0 &  83.3\\
\hline
			
			\hline
		\end{tabular}
	}
	
	\caption{Comparison of GLUE dev and test sets. Our model is in boldface. MT-DNN dev results are from  \cite{liu-etal-2019-multi} and other dev results are from \cite{anonymous2020electra}.}\label{GLUE dev set results}
	
\end{table*}

\begin{table}[t!]
 		\resizebox{\linewidth}{!}
	{
	\begin{tabular}{l c c c}%{p{2.5cm} p{1cm} p{1cm} p{1cm} p{1cm}}
		\hline
		
		\hline
		\multirow{2}{*}{System} &  GLUE  &  SNLI & SNLI \\ 
		 &   Dev  &  Dev & Test \\ 
		\hline
% 		\multicolumn{3}{c}{\emph{In literature}} \\
		\textbf{LIMIT-BERT}  & 82.6 & \bf90.6 & \bf91.0\\
		\hline
		\quad w/o Multi-Task  & \bf 82.9 & 90.5 & 90.7 \\
% 		\quad $\delta$ & - & 91.3 & - & 91.3 \\
		\quad w/o ELECTRA & 81.3 & 90.4 & 90.5 \\
% 		\quad $\delta$ & - & 91.3 & - & 91.3 \\
		\quad w/o SPM & 80.2 & 90.6 & 90.8 \\
% 		\quad $\delta$ & - & 91.3 & - & 91.3 \\
		
		\hline
	\end{tabular}
}
	{
		\caption{\label{tab:ablation_mrc} Ablation study of LIMIT-BERT (base) on GLUE and SNLI.} } 
\end{table}

% \begin{table}[t!]
%     \small
% 		\resizebox{\linewidth}{!}
% 	{
% 	\begin{tabular}{l c c c c}%{p{2.5cm} p{1cm} p{1cm} p{1cm} p{1cm}}
% 		\hline
		
% 		\hline
% 		\textbf{TASK} &  \textbf{Synconst } & \textbf{Syndep } & \textbf{SRLspan} & \textbf{SRLdep} \\ 
% 		\textbf{} &  \textbf{ Dev} & \textbf{ Test} &  \textbf{ Dev} & \textbf{ Test} \\ 
% 		\hline
% 		\hline
% % 		\multicolumn{3}{c}{\emph{In literature}} \\
% 		\textbf{LIMIT-BERT} &82.6 & 90.1 &- & 90.1\\
% 		\hline
% 		\quad - Linguistics-Task & - & 91.3 & - & 91.3 \\
% % 		\quad $\delta$ & - & 91.3 & - & 91.3 \\
% 		\quad - ELECTRA & - & 91.3 & - & 91.3 \\
% % 		\quad $\delta$ & - & 91.3 & - & 91.3 \\
% 		\quad - SPM & - & 91.3 & - & 91.3 \\
% % 		\quad $\delta$ & - & 91.3 & - & 91.3 \\
		
% 		\hline
% 	\end{tabular}
% }
% 	{
% 		\caption{\label{tab:ablation_lig} Ablation study of LIMIT-BERT with BERT$_\text{base}$.} } 
% \end{table}

\begin{table}[t!]
    % \small
	\centering
	\resizebox{\linewidth}{!}{
	\begin{tabular}{lccccc}  
		\toprule  
		\multirow{2}{*}{System}
		&\multicolumn{1}{c}{SEM$_{span}$}
		&\multicolumn{1}{c}{SEM$_{dep}$}
		&\multicolumn{1}{c}{SYN$_{con}$}
		&\multicolumn{2}{c}{SYN$_{dep}$}
		\cr  
		\cmidrule(lr){2-2} 
		\cmidrule(lr){3-3} 
		\cmidrule(lr){4-4}
		\cmidrule(lr){5-6}
		 &F$_1$&F$_1$&F$_1$&UAS&LAS\cr
	\midrule
	\textbf{LIMIT-BERT}  &\bf86.25 &\bf85.74 &95.34 &96.59 &\bf94.71\cr
	\midrule
	\quad w/o Multi-Task &85.60 &85.24 &95.16 &96.38 &94.38\cr
    \quad w/o ELECTRA &86.20 &85.71 &95.32 &96.59 &94.70\cr
    \quad w/o SPM &86.21 &85.72 &\bf95.44 &\bf96.64 &94.70\cr
	\bottomrule  
	\end{tabular}
	}
	\caption{\label{tab:ablation_lig} Ablation study of LIMIT-BERT (base) on linguistics tasks.}
\end{table}

\subsection{Discussions} 

\noindent \textbf{Ablation Study} \quad
LIMIT-BERT contains three key components: Multi-Task learning, ELECTRA training approach, and Syntactic/Semantic Phrase Masking (SPM).
To evaluate the contribution of each component in LIMIT-BERT, we remove each component from the model for training and then fine-tune on downstream NLU tasks and  linguistics tasks for evaluation. 
In consideration of computational cost, we apply BERT$_\text{base}$ as the start of training and only use one-tenth of the BERT training corpus.
We employ the same training setting for each ablation model: 200k training steps, 1e-5 learning rate and 32 batch size.
After language model training, we extract the layers of BERT$_\text{base}$ and fine-tune on downstream tasks for evaluation.

The ablation study is conducted on NLU tasks and linguistics tasks shown in Table \ref{tab:ablation_mrc} and \ref{tab:ablation_lig} respectively.
For NLU tasks, GLUE and SNLI results both decrease in w/o ELECTRA and w/o SPM setting showing the effectiveness of the ELECTRA training approach and SPM for NLU tasks. 
For linguistics tasks, w/o Multi-Task setting hurts the performance obviously from LIMIT-BERT both of semantic and syntactic parsing, which shows the effectiveness of multi-tasks learning for linguistics tasks. 
Besides, the ELECTRA training approach and SPM also can improve performance when fine-tuning on linguistics tasks. 

Comparing the results in Tables \ref{tab:ablation_mrc} and \ref{tab:ablation_lig}, ELECTRA training approach and SPM are more effective for NLU tasks while multi-tasks learning can improve the linguistics tasks performance significantly.
The possible explanation is that multi-tasks learning enables LIMIT-BERT to 'remember' the linguistics information and thus lead to better performance in downstream linguistics tasks.
% While SPM can 

% We examine the fine-tuning effect of LIMIT-BERT on linguistics tasks.
% The results in Table \ref{Fine-tuning effect} show that LIMIT-BERT with or without finetuning still outperforms BERT$_\text{WWM}$ baseline consistently among all tasks.
% In such a case, fine-tuning is necessary to boost
% the semantic parsing performance while no-fine-tuning performs better on syntactic parsing. As shown in Table \ref{Fine-tuning effect}, the accuracy
% improves 0.1 F1 and 0.4 F1 of span SRL and dependency SRL after fine-tuning respectively but no-fine-tuning performs better nearly 0.2 F1 of syntactic parsing.

\noindent \textbf{Fine-tuning Effect} \quad
We examine the fine-tuning effect of LIMIT-BERT on linguistics tasks.
The results in Table \ref{Fine-tuning effect} show that LIMIT-BERT with or without finetuning still outperforms BERT$_\text{WWM}$ baseline consistently among all tasks.
In such a case, fine-tuning is necessary to boost
the semantic parsing performance while no-fine-tuning performs better on syntactic parsing. As shown in Table \ref{Fine-tuning effect}, the accuracy
improves 0.1 F1 and 0.4 F1 of span SRL and dependency SRL after fine-tuning respectively but no-fine-tuning performs better nearly 0.2 F1 of syntactic parsing.
The possible explanation is that no-fine-tuning LIMIT-BERT use semi-supervised training data which contains much more long sentence samples and benefits syntactic parsing more.

\begin{table}[t!]
    % \small
	\centering
	\resizebox{\linewidth}{!}{
	\begin{tabular}{lccccc}  
		\toprule  
		\multirow{2}{*}{System}
		&\multicolumn{1}{c}{SEM$_{span}$}
		&\multicolumn{1}{c}{SEM$_{dep}$}
		&\multicolumn{1}{c}{SYN$_{con}$}
		&\multicolumn{2}{c}{SYN$_{dep}$}
		\cr  
		\cmidrule(lr){2-2} 
		\cmidrule(lr){3-3} 
		\cmidrule(lr){4-4}
		\cmidrule(lr){5-6}
		 &F$_1$&F$_1$&F$_1$&UAS&LAS\cr
	\midrule
	Baseline  &86.55 &86.10 &95.52 &96.54 &94.71\cr
	LIMIT-BERT &\bf87.13 &\bf86.77 &95.55 &96.54 &94.74\cr
    LIMIT-BERT$\dagger$ &87.04 &86.38 &\bf95.72 &\bf96.82 &\bf94.82\cr
	\bottomrule  
	\end{tabular}
	}
	\caption{Fine-tuning effect analysis on English dev sets, no finetuning result is marked by $\dagger$.}\label{Fine-tuning effect}
\end{table}

\begin{table}[t!]
\small
		\resizebox{\linewidth}{!}
	{
	\begin{tabular}{p{5cm} p{1cm}}
		\hline
		\textbf{Model} & \textbf{Test}   \\ 
		\hline
% 		\multicolumn{3}{c}{\emph{In literature}} \\
		\citet{yasunaga-etal-2018-robust}  & 97.59 \\
		\citet{akbik-etal-2018-contextual}  & 97.85\\
		\citet{bohnet-etal-2018-morphosyntactic}   & 97.96\\
		\textbf{LIMIT-BERT}  & 97.71 \\
		\hline
		
	\end{tabular}
}
	{
		\caption{\label{tab:pos} POS tagging accuracy on the WSJ test set, with other top-performing systems.} } 
	
\end{table}

\begin{figure}[t!]
    \centering
    \includegraphics[width=3in]{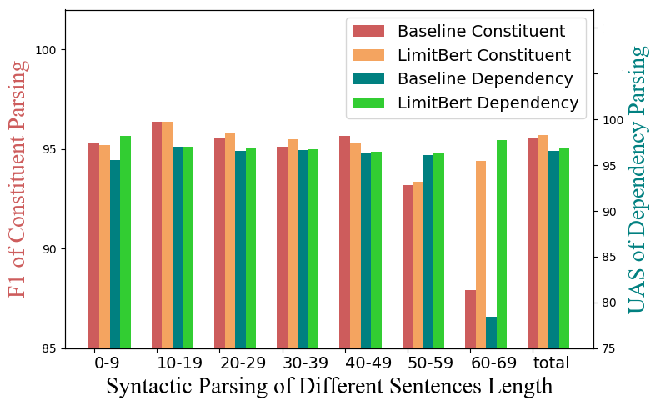}
    \includegraphics[width=3in]{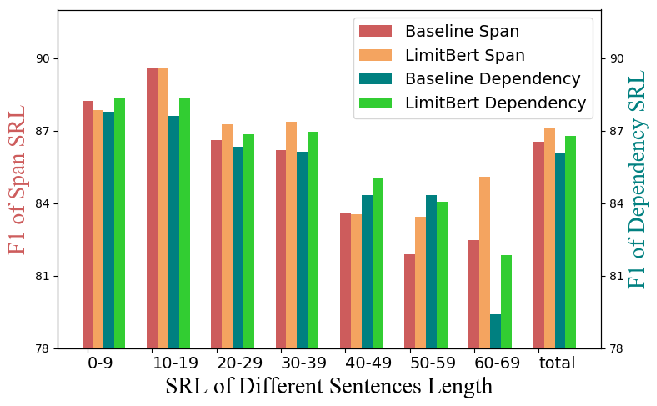}
    \caption{ The performance of baseline model and LIMIT-BERT while varying sentence length of four linguistics tasks on the English dev set.}
    \label{fig4}
\end{figure}

\noindent \textbf{Part-Of-Speech Performance} \quad
Table \ref{tab:pos} lists the results of POS tagging on WSJ test set showing that our LIMIT-BERT achieves competitive results compared with other state-of-the-art models.
Note that we only apply simple one-layer decoder without those complicated ones such as conditional random field (CRF) \cite{Lafferty-2001} as the POS tagging task is not the main concern of our model.

\noindent \textbf{Sentences Length Performance} \quad
The performance of baseline model and LIMIT-BERT while varying the sentence length of four linguistics tasks on the English dev set is shown in Figure \ref{fig4}. 
The statistics show that our LIMIT-BERT outperforms the baseline model of over all sentence lengths.
% Note that
% the subset is small (only 391 sentences), so the effect may be random.
For different sentence lengths, LIMIT-BERT outperforms much better than baseline model on long sentence (larger than 50) of both syntactic and semantic parsing. 
The possible explanation is that LIMIT-BERT uses semi-supervised training data, which contains much more long sentence samples and benefits parsing performance on long sentences.
% on sentences of between 50 and 59 words. 
% Note that
% the subset is small (only 391 sentences), so the effect may be random.

\section{Related Work}

% \subsection{Linguistics Inspired NLP Tasks}
\noindent \textbf{Linguistics Inspired NLP Tasks} \quad
With the impressive success of deep neural networks
in various NLP tasks \cite{ChenD14, Dozat2017Deep, Ma2018Stack,strubell-etal-2018-linguistically, luo-zhao-2020-bipartite,Li2020Data-dependent, he-etal-2019-syntax, luo2019named,zhang-etal-2018-exploring,li-etal-2018-seq2seq, zhang-etal-2018-modeling}, syntactic parsing and semantic role labeling have been well developed with neural network and achieve very high performance \cite{ChenD14, Dozat2017Deep, Ma2018Stack, Kitaev-2018-SelfAttentive, zhou-zhao-2019-head}. 
Semantic role labeling is deeply related to syntactic structure and a number of works try to incorporate syntactic information in semantic role labeling models by different methods such as concatenation of lexicalized embedding \cite{he-etal-2018-syntax}, usage of syntactic GCN \cite{li-etal-2018-unified} and multi-task learning \cite{strubell-etal-2018-linguistically, zhou2019parsing}.
Besides semantic role labeling and syntactic parsing are two key tasks of semantics and syntax so that they are included into our linguistics tasks for multi-task learning.

In addition, both span and dependency are popularly adopted annotation styles for both semantics and syntax and some work on jointly learning of semantic and syntactic \cite{henderson-etal-2013-multilingual, lluis-etal-2013-joint, swayamdipta-etal-2016-greedy} .
Researchers are interested in two styles of SRL models that may benefit from each other rather than their separated development, which has been roughly discussed in \cite{johansson-nugues-2008-dependency}.
On the other hand, researchers have discussed how to encode lexical dependencies in phrase structures, like lexicalized tree adjoining grammar (LTAG) \cite{SCHABESC88-2121}
, Combinatory Categorial Grammar (CCG) \cite{steedman2000syntactic} 
and head-driven phrase structure grammar (HPSG) \cite{pollard1994head} which is a constraint-based highly lexicalized non-derivational generative grammar framework. 
To absorb both strengths of span and dependency structure, we apply both span (constituent) and dependency representations of semantic role labeling and syntactic parsing.
Thus, it is a natural idea to study the relationship between constituent and dependency structures, and the joint learning of constituent and dependency syntactic parsing \cite{KleinP04,CharniakP05-1022,FarkasW11-2924,GreenW12-0503,RenCombine2013,XuP14-1021,YoshikawaP17-1026}.

% \subsection{Pre-trained Language Modeling}
% Recently, much progress has been made in general-purpose language modeling that can be used across a wide range of tasks (Radford et al. 2018; Devlin et al. 2018; Zhang et al. 2020b; Zhou, Zhang, and Zhao 2019; Zhang et al. 2019).
% Obviously, it requires a good representation of the meaning of a sentence.
\noindent \textbf{Pre-trained Language Modeling} \quad
Recently, deep contextual language model has been shown effective for learning universal language representations by leveraging large amounts of unlabeled data, achieving various state-of-the-art results in a series of NLU benchmarks. Some prominent examples are Embedding from Language models (ELMo) \cite{PetersN18-1202}, Bidirectional Encoder Representations from Transformers (BERT) \cite{Jacobbert} and Generalized Autoregressive Pretraining (XLNet) \cite{XLNet-Zhilin-2019}.

Many latest works make attempts to modify the language model based on BERT such as ELECTRA \cite{anonymous2020electra} and MT-DNN \cite{liu-etal-2019-multi}.
ELECTRA focuses on the [MASK] tokens mismatch problem and thus combines the idea of 
Generative Adversarial Networks GANs \cite{goodfellow2014generative}.
MT-DNN applies multi-task learning to language
model pre-training and achieves  new state-of-the-art results on GLUE benchmark.
Besides, \cite{gururangan-etal-2020-dont} finds that multiphase adaptive pretraining offers large gains in task performance which is similar with our semi-supervised learning strategy.

% Multi-Task Learning (MTL) is inspired by human
% learning activities where people often apply
% the knowledge learned from previous tasks to help
% learn a new task (Caruana, 1997; Zhang and Yang,
% 2017).

\section{Conclusions}

In this work, we present LIMIT-BERT which applies multi-task learning with multiple  linguistic  tasks by semi-supervised  learning.
We use five key syntax and semantics tasks
:   Part-Of-Speech  (POS)  tags,  constituent  and  dependency  syntactic  parsing,  span  and  dependency semantic role labeling (SRL).
and further improve the masking strategy of BERT training by effectively exploiting the available syntactic and semantic clues for language model training.
The experiments show that LIMIT-BERT 
% obtains new state-of-the-art or competitive results 
outperforms the strong baseline BERT$_\text{WWM}$ 
on four benchmark parsing treebanks and two NLU tasks.
The results of GLUE and SNLI empirically illustrate that incorporating  linguistic  knowledge  into  pre-training language  BERT by multi-task and semi-supervised  learning  can also enhance downstream tasks.
There are many future areas to explore to improve
LIMIT-BERT, including a deeper understanding
of model structure sharing in MTL, a more
effective training method that leverages relatedness
among multiple tasks, for both fine-tuning
and pre-training, and ways of
incorporating the linguistic structure of text in a
more explicit and controllable manner.

\bibliography{anthology,emnlp2020}
\bibliographystyle{acl_natbib}

\end{document}